\documentclass[letterpaper, 10 pt, conference]{ieeeconf}  

\IEEEoverridecommandlockouts                              

\usepackage{graphicx} 
\usepackage{amsmath} 
\usepackage{amssymb} 
\usepackage{algorithm}
\usepackage{gensymb}
\usepackage[noend]{algpseudocode}

\title{\LARGE \bf
An Approach to Autonomous Science by Modeling Geological Knowledge in a Bayesian Framework
}
\date{}

\author{Akash Arora$^{1}$, Robert Fitch$^{1,2}$ and Salah Sukkarieh$^{1}$
\thanks{*This work was supported by the Australian Centre for Field Robotics}
\thanks{{$^{1}$Australian Centre for Field Robotics, The University of Sydney, NSW, Australia}
{$^{2}$Centre for Autonomous Systems, University of Technology Sydney, NSW, Australia}
{\tt\small a.arora@acfr.usyd.edu.au}
}
}

\begin{document}

\maketitle
\thispagestyle{empty}
\pagestyle{empty}

\begin{abstract}
\emph{Autonomous Science} is a field of study which aims to extend the autonomy of exploration robots from low level functionality, such as on-board perception and obstacle avoidance, to science autonomy, which allows scientists to specify missions at task level. This will enable more remote and extreme environments such as deep ocean and other planets to be studied, leading to significant science discoveries. This paper presents an approach to extend the high level autonomy of robots by enabling them to model and reason about scientific knowledge on-board. We achieve this by using Bayesian networks to encode scientific knowledge and adapting Monte Carlo Tree Search techniques to reason about the network and plan informative sensing actions. The resulting knowledge representation and reasoning framework is anytime, handles large state spaces and robust to uncertainty making it highly applicable to field robotics. We apply the approach to a Mars exploration mission in which the robot is required to plan paths and decide when to use its sensing modalities to study a scientific latent variable of interest. Extensive simulation results show that our approach has significant performance benefits over alternative methods. We also demonstrate the practicality of our approach in an analog Martian environment where our experimental rover, Continuum, plans and executes a science mission autonomously.  
\end{abstract}

\section{Introduction}
Information gathering using mobile robots in dangerous or hard-to-access environments has significantly improved humanity's ability to understand our world~\cite{Cliff-RSS-15,dunbabin2012robots}. Research in improving the capabilities of these robots has largely focused on automating low level functionality, such as perception and obstacle avoidance. Higher level reasoning (and task level autonomy in particular) in unstructured real world environments has not received as much attention. However, this technology is critical to enable the study of more remote areas, where much of the interesting science lies. Such high level autonomy in the context of information gathering missions is known as \emph{Autonomous Science}. In this paper we use robotic planetary exploration as our motivating application, but the ideas presented here are applicable to exploration of remote environments in general.

\begin{figure}[t]
\centering
\includegraphics[width=0.45\textwidth]{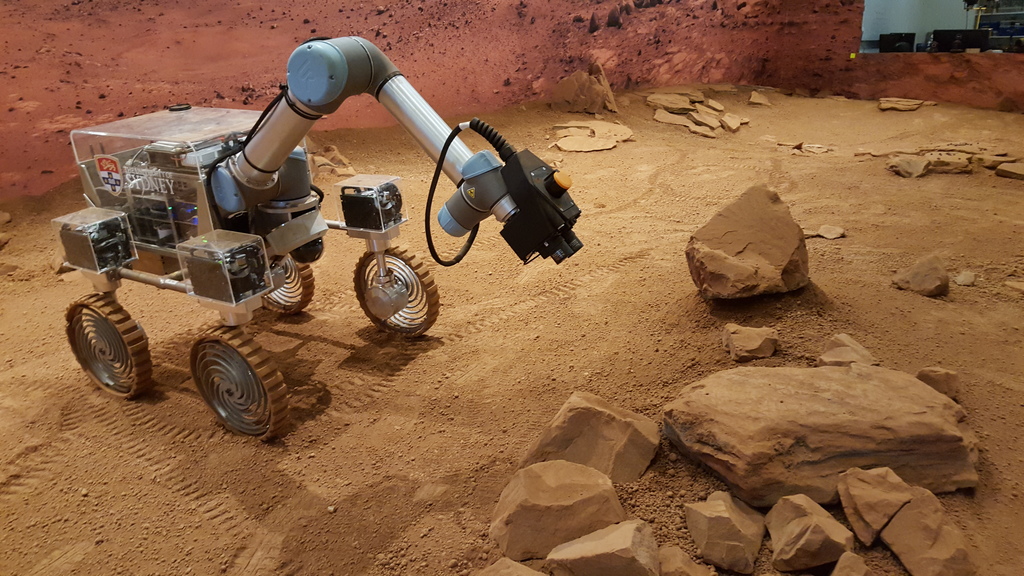}
\caption{The Continuum rover in a Mars-analog environment using its robotic arm to closely examine rocks}
\label{fig:continuum}
\end{figure}

Planetary rovers are required to explore largely unknown environments under strong communication constraints such as high latency, limited bandwidth and infrequent communication windows. They are equipped with multiple heterogeneous sensors which must be used collaboratively to achieve a set of high level scientific goals such as finding evidence of water. In outdoor environments there is also significant noise in the form of shadows, sensor inaccuracies, and deformable  terrain. These challenges induce the need for some form of autonomy to ensure safety and mission progress in the absence of human supervision.

Recent research in Autonomous Science has explored increasing autonomy through anomaly detection, selective data transmission, guiding data collection with template based feature matching and adaptive sampling through non-parametric models such as Gaussian processes (GPs) \cite{castano2007oasis,thompson2011autonomous,woods2009autonomous}.
Higher level reasoning such as deciding where to go in the short and long term, which sensors to deploy and most importantly, making inferences from observations to update scientific hypotheses, is handled primarily by human supervisors on Earth. This creates a bottleneck in the scientific progress made as communication can typically only be established twice a day on Mars. In this work we approach the problem of Autonomous Science from a novel cognitive robotics perspective by equipping the rover with an approximate representation of a scientist's domain knowledge. We then develop techniques to reason about this knowledge to explore and sample the environment in a more intelligent and goal-driven manner.  

We represent geological knowledge as a Bayesian network~(BN). The BN's structure and conditional probability parameters allow us to capture many important aspects of scientific knowledge such as conditional dependencies between variables, causality relationships and any mathematical or process models that may be known prior to the mission. BNs are limited in expressivity as compared to knowledge representation languages such as Answer Set Prolog \cite{zhang2015mixed}, but have the advantage of handling uncertainty more robustly. This property is crucial in unstructured environments, such as Mars, where sensors and controls are both noisy. Further, there are many algorithms which allow fast approximate inference in BNs, which is an important advantage lacking in many other languages \cite{pearl2014probabilistic}. 

We then show how Monte Carlo Tree Search~(MCTS) techniques can be applied to reason about the knowledge BN efficiently and plan goal oriented sensing actions over long horizons. The resulting knowledge representation and reasoning framework extends robotic information gathering in two ways: it enables the robot to reason about prior scientific knowledge in a principled manner, and it allows the robot to plan with multiple sensing modalities to study latent environmental variables which cannot directly be observed. 

We apply the framework to a Mars exploration mission where the robot observes environmental features to determine the geological identity of different regions on the map, such as ancient riverbeds and volcanic zones. The robot is equipped with two sensors, a camera and an idealized spectrometer and required to autonomously plan where to move and which sensor to use at each time step while satisfying some sensing budget. We present extensive simulation results where our method outperforms alternative approaches in terms of information gain (confidence) and accuracy. We conclude by demonstrating the practicality of our approach in an analog Martian environment using our experimental rover, Continuum (shown in Fig.~\ref{fig:continuum}).  

\section{Related Work}
\subsection{Bayesian Networks for Knowledge Representation}
Due to their desirable property of remaining robust under uncertainty, many authors have employed BNs to model domain knowledge, particularly in the form of expert systems. Applying these networks to robotic decision making problems in unstructured environments is, however, less studied. Most authors have limited their use to classification and have not closed the loop around path planning~\cite{sharif2015autonomous,apostolopoulos2001robotic}.  

Work that is similar to ours is by Post et al.~\cite{post2016planetary}, who use BNs to create an obstacle map while integrating any sensor uncertainties that are present. A path is then planned to achieve a goal position while minimizing the probability of collisions. This work, however, does not attempt to model scientific knowledge, especially the spatial relationships.

Gallant et al.~\cite{gallant2011science} used a BN to classify minerals and assign benefit scores based on the current scientific goals of the mission. The benefit scores were then fed into a cost function to determine the best action take. However, their approach does not reason about unobserved parts of the environment and does not consider the problem of selecting which sensor to use. 

\subsection{Informative Path Planning}
The idea of planning the placement of sensors to achieve some information-theoretic goal can be viewed as an active sensing problem, or more generally, an informative path planning problem. When the problem is monotone submodular, greedy approaches are effective and offer performance guarantees~\cite{krause2012near}. Unfortunately, this property is often violated in field environments leading to arbitrarily poor worst case performance. Branch and bound techniques which prune suboptimal branches early in the tree search have shown promise~\cite{bestprobabilistic}. However efficiently calculating tight bounds in problems with unknown environments and multiple sensors like ours is non-trivial. MCTS methods, however, work for any general objective function and do not require bounds. They are anytime and hence suitable for online planning~\cite{browne2012survey}. 

Approaches that involve initially unknown environments typically utilize GPs and exploit the monotone submodular nature of the mutual information or variance reduction function to avoid exhaustive search~\cite{binney2012branch,hollinger2014sampling}. While GPs can represent spatial phenomena in a probabilistic manner, they are not particularly useful tools for encoding domain knowledge especially causal knowledge. Proposed methods are limited to: imposing priors on the co-variance parameters, transforming the training data and biasing the mean function~\cite{azmanincorporating}. Further, the computational complexity of GPs make them difficult to use in online planning applications with long horizons such as the problem considered here.  

\section{Autonomous Science for Planetary Rovers}
This section discusses the robot properties, the assumptions made about the world, and formally defines the planning problem that the robot is required to solve in the context of Mars exploration.

\subsection{Robot and Environment Setup}
The robot is a UGV which moves around in a world discretized into cells. The robot is equipped with a camera which can detect rocks and extract their visual features. The camera can take measurements within its field of view which may span multiple cells. The robot is also equipped with an ultraviolet (UV) light source which it can project onto the environment to reveal UV reflective minerals. The UV light source simulates what a spectrometer might do on a real Mars mission since it is energetically expensive to use and has a narrow sensing range, but gives more informative measurements than a camera. For the remainder of this paper we refer to the camera as the low cost `remote' sensor and the UV source as the high cost `local' sensor.

\subsection{Problem Setup}
Given this robot and environment setup and some representation of scientific knowledge, the robot is required to plan a sequence of informative sensing actions $a_{seq}$ to minimize entropy of some scientific latent variable of interest $L$ across all of the $N$ cells on the map. A sensing action is a tuple consisting of a movement action and which sensor to use. The robot is also constrained to some specified sensing (energy or time) budget. The optimization objective can be described by Eq. 1. 

\begin{equation}
\begin{split}
&a^*_{seq} = \operatorname*{arg\,max}_{a_{seq} \in A} EI(a_{seq}) \\
&\textbf{s.t.} \sum^{|a_{seq}|}_{i}{\textnormal{cost($a_i$)} = \text{Budget}}
\end{split}
\end{equation}

The cost function and the action space $A$ we use will be defined in Section V. $EI$ is the expected information gain of an action sequence which is calculated by marginalizing out all possible observations $Z_{seq}$ that can result from the sensing sequence (Eq. 2). The $P(Z_{seq}|a_{seq})$ term is effectively a sensor model and $I$ is an information gain function given by Eq. 3 where $H$ is the Shannon entropy. The conditional entropy $H(L_n|Z_{seq})$ requires a mapping from observations to the latent variable of interest. This is the knowledge representation component of the framework while the optimization to determine sensing sequences is akin to scientific reasoning. 

\begin{equation}
EI(a_{seq}) = \sum_{Z_{seq}}{I(Z_{seq})P(Z_{seq}|a_{seq})}
\end{equation}

\begin{equation}
I(Z_{seq}) = \sum_{n=1}^{N}{H(L_n) - H(L_n|Z_{seq})}
\end{equation}

\section{Approach}
In this section we present the two main components of the system: BN knowledge representation and a MCTS planner. The planner reasons about the knowledge network and the robot and environment state to determine a sequence of sensing actions which maximize the information gained on the scientific latent variable of interest. 

\subsection{Knowledge Representation}
The purpose of the BN is to model the relationship between the observations made and the latent variable of interest through scientific knowledge. The structure of the network encodes causal knowledge while the conditional probability parameters encode quantitative knowledge. 

\begin{figure}[!t]
\centering
\includegraphics[trim={5cm 3cm 8cm 6.5cm},clip,width=0.5\textwidth]{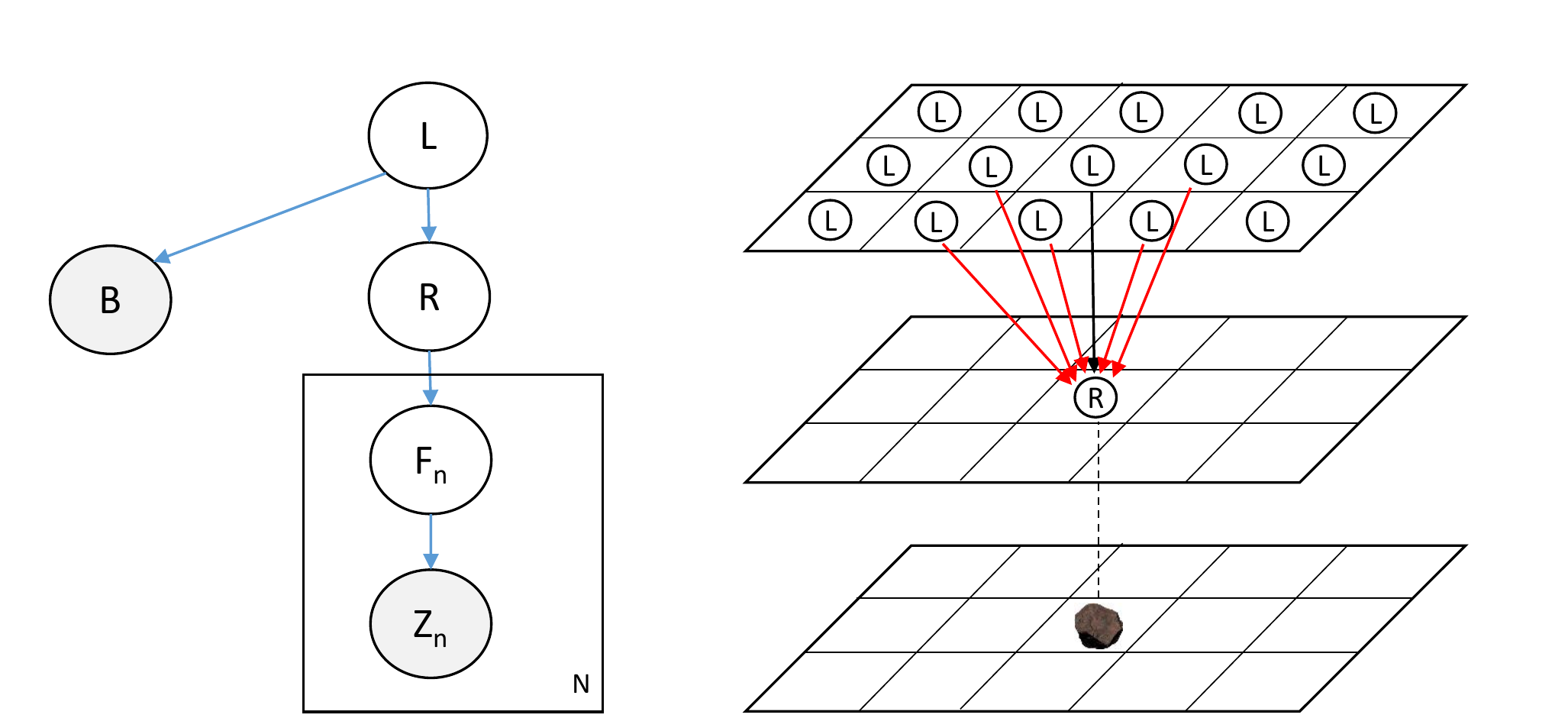}
\caption{Left: The structure of the Bayesian network used to represent geological knowledge. Right: Spatial relationships between adjacent cells}
\label{knowledgebn}
\end{figure}

Since, rocks are key sources of geological cues in Martian environments, we design the BN around them (Fig.~\ref{knowledgebn}). The rocks in the environment of class $R$ exhibit $N$ visual features represented by the variable $F$. The robot can observe these features through its camera, denoted by the variable $Z$. The variable $B$ is the UV reflective material that can be measured by the robot's local sensor. Lastly $L$ resembles the underlying latent variable which affects the environment. In this paper we assume $L$ to be the type of location the robot is in such as desert or a riverbed and this is scientifically interesting variable we are interested in gaining information on.  

All nodes in the network are discrete as geologists often look for features which do not have associated continuous measurements such as the presence of bedding on a rock. Discretization also simplifies inference. The structure of the BN can be adapted to account for different variables and dependencies that come with specific applications. In this paper all nodes have three categories they can take but the approach works for any arbitrary number. 

The proposed BN structure allows several sources of information to be integrated in the form of conditional probabilities. $P(Z|F)$ is the sensor model, $P(F|R)$ is effectively a classifier likelihood while $P(B|L)$ and $P(R|L)$ are geological properties of the environment. This network exists in every cell of the environment. If there are no rocks detected in a cell, then the $R$ node and its children will be removed to speed up future computations. 

In natural environments there are also strong spatial correlations present. There are several methods of encoding this relationship. A common approach is through a Markov random field. However this will make the inference problem difficult as cycles will be introduced in the graphical model. Another alternative is to add links between the $L$, $R$ and $B$ nodes of adjacent cells. This implies that the variables $R$ and $B$ are dependent on the $L$ nodes in the neighborhood as opposed to just the one in its cell. Nodes that are far from where the observation was taken have less influence on the inference. This decreasing influence is modeled by a Gaussian function. Fig. \ref{knowledgebn} illustrates this spatial dependency.  The resolution of the $L$ grid does not have to match the $R$ grid and can be adapted based on the expected spatial variability of variables. 

The conditional probability parameters can either be specified directly through domain knowledge, learned from training data \cite{heckerman1995learning} or even learned online by modeling them as Dirichlet distributions \cite{girdhar2015modeling}. In this work we assume the maximum likelihood parameters are known a priori. 

Due to this BN's structure, the belief on the value of nodes can be updated recursively without keeping an history of observations. The message passing technique is used for efficiently propagating belief updates through the BN \cite{yedidia2000generalized}.

\subsection{Monte Carlo Tree Search}
In this problem, the robot acquires observations after executing every sensing action and has the freedom to adapt the sensing plan accordingly. Therefore at planning time, the robot only needs to decide the next best action to take which in expectation will give maximal future rewards. We propose the use of MCTS methods to address this sequential decision making problem. The algorithm is presented in Alg. 1. 

MCTS is a best first, anytime algorithm which involves cycling through four stages: node selection, expansion, simulation and back-propagation. The key idea is to first select promising leaf nodes based on a tree policy. The selected node is expanded and a terminal reward is estimated by conducting simulations or `rollouts' in the decision space. The reward is then back propagated up the tree and the process is repeated until some computational budget is reached. At the end of the search, the child of the root node with the highest average reward is selected as the next best action. Since MCTS is sampling based, it is well suited for large state spaces, high branching factors and long horizon planning. For an overview on MCTS methods we refer the reader to Browne's comprehensive survey \cite{browne2012survey}. 

\begin{algorithm}[!t]
\caption{MCTS Science Autonomy Planner}
\begin{algorithmic}[1]
\State \textbf{Input:} SensingBudget $S$, BeliefSpace $Bel$, DomainKnowledge BN $K$, RemainingBudget $R$
\Function{Main}{} 
	\State $R \gets S$
	\While{$R > 0$}
		\State $robotPose \gets getLocalisation()$
        \State $a_{opt} \gets planner(robotPose, R, Bel, K)$
        \State $Z\gets takeObservation(a_{opt})$
		\State $Bel\gets updateBeliefSpace(Z, Bel, K)$
        \State $R \gets R - cost(a_{opt})$
	\EndWhile
\EndFunction
\State
\Function{planner}{$robotPose, R, Bel, K$}
    \State $T \gets initialiseTree(robotPose, R)$
	\State $currentNode \gets T.rootNode$
	\While{within computational budget}
		\State $currentNode \gets treePolicy(T)$
		\State $sequence\gets rolloutPolicy(currentNode, R)$
        \State $reward \gets getReward(sequence, Bel, K)$
        \State $T \gets updateTree(T, reward)$
	\EndWhile
    \State
    \Return $bestChild(T)$
\EndFunction
\State
\Function{rolloutPolicy}{$currentNode, R$}
	\State $sequence \gets currentNode$
    \While{$R > 0$}
    	\State $nextNode \gets defaultPolicy(currentNode)$
        \State $currentNode \gets nextNode$
        \State $sequence \gets sequence + currentNode$
        \State $R \gets currentNode.R$
    \EndWhile
    \State
    \Return $sequence$
\EndFunction
\State
\Function{getReward}{$sequence, B, K$}
	\State $reward \gets 0$
    \For{$i=1:length(sequence)$}
    	\State $currentAction \gets sequence(i)$
        \State $Z = sampleObs(currentAction, Bel, K)$
        \State $Bel_{new} = updateBelief(Z, Bel, K)$
        \State $infoGain = calcInfoGain(Bel_{new}, Bel)$
        \State $reward \gets reward + infoGain$
        \State $Bel \gets Bel_{new}$
    \EndFor
    \State
    \Return $reward$

\EndFunction

\end{algorithmic}
\end{algorithm}

We formulate the MCTS such that each node in the tree is a potential sensing action that can be made. It is a tuple consisting of the robot's x and y position, the orientation, the type of sensor used and the remaining sensing budget. Each node also stores the average reward $\bar{R_i}$ of all the simulations that have passed through it and the number of times it has been visited $n_i$ during the tree search. The children of the node are determined by the robot's action space and the remaining budget. We now describe each stage of the MCTS in detail and show how it has been adapted for our problem. 

\textbf{Selection:} The first stage of MCTS is using a tree policy to select which leaf nodes to expand. We want to expand leaf nodes which are expected to have a good terminal reward but at the same time evaluate alternative nodes sufficiently to minimize chances of converging to local minima. The Upper Confidence Tree (UCT) policy based on the optimism in the face of uncertainty paradigm is known to be a good solution to balance the exploration/exploitation trade-off present here \cite{kocsis2006bandit}. UCT begins at the root node and iteratively selects leaf nodes with the highest Upper Confidence Bound (UCB) until a node with unexpanded children is reached. The UCB score for node $i$ is defined by Eq. 4 below. 

\begin{equation}
UCB_i = \bar{R_i} + C_p\sqrt\frac{2\log N}{n_i}
\end{equation}

The first term is the `exploitation' component of UCB. $\bar{R_i}$ is the average reward of all rollouts that have passed through $node_i$. We define the reward function in the simulation subsection below. The second term in the equation is the `exploration' component where $N$ is the number of times the parent of the node has been evaluated and $n_i$ is the number of times node $i$ has been evaluated. $C_p$ is a constant that balances exploration and exploitation. We found empirically that a value of $0.1$ gave good results in both simulations and hardware experiments. 

\textbf{Expansion:} From the leaf node selected by the UCT policy, an unexpanded child node is randomly selected and added to the tree. 

\textbf{Simulation:} The aim of the simulation stage is to determine the terminal reward associated with this newly expanded child node by executing some default policy. Here we use a random action selection policy from the selected node until the sensing budget is exhausted. A random policy was used because it requires minimal computational overhead to calculate and ensures the decision space is uniformly explored. However since we are sampling randomly, a large number of rollouts are often required to accurately estimate rewards. Using problem specific rollout policies has been shown to significantly improve tree convergence but we leave this as an interesting avenue for future work. 

The expected information gain function defined earlier in Eq. 2 is the ideal reward function to evaluate a rollout. However, calculating this function analytically requires summing over all possible observations that can result from the rollout sequence. In our problem, the low cost remote sensor observes rocks in its field of view. Each rock can exhibit $|F|^N$ combinations of features where $|F|$ is the number of classes each feature can take and $N$ is the total number of features. Furthermore the number of rocks seen as well as the position of the rocks in the image are all unknown at planning time if an area hasn't been observed before. The observation space is therefore very large and evaluating the reward exactly is not practical. 

We define the reward function as $\frac{I_r}{H_{init}}$ where $I_r$ is the information gain during rollout $r$ and $H_{init}$ is the joint entropy of the $L$ variables at the current state of the mission. This division constrains the average reward to between $0$ and $1$- a requirement for UCB convergence guarantees to hold. We approximate $I_r$ by sampling. We begin at the first node of the rollout sequence. Depending on the sensing action used, an observation is sampled from the belief space. The belief space is updated and passed onto the next node. The process is iterated until the terminal node is reached. The total information gain is determined by subtracting the entropies of the initial and terminal belief spaces. 

\textbf{Back-propagation:} Lastly the reward received by the rollout is back-propagated up the tree and the average reward and number of evaluations for each node involved is updated. 

The four stages are repeated until the computational budget for the robot has expired at which point the root child with the highest average reward is selected as the next best action. Given enough samples and an appropriate value for the exploration parameter $C_p$ in Equation 4 it can be shown that the tree will converge to the optimal action sequence.

\section{Simulation Experiments}

\begin{figure}[t]
\centering
\includegraphics[trim = {0 0.2cm 0 0},clip,width = 0.5\textwidth]{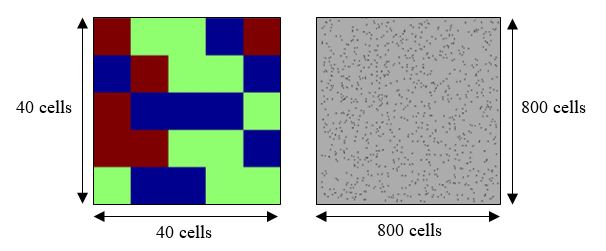}
\caption{Left: An example of a randomly generated location ground truth map where the colors signify different classes. Right: An example rock map generated by sampling from the Bayesian Network}
\label{egmap}
\end{figure}

This section aims to empirically demonstrate the performance of the MCTS planner for our Autonomous Science problem. While there are several algorithms for informative path planning in literature, they cannot be applied in situations where multiple sensing modalities are involved without significant algorithmic modifications and heuristics. We therefore compare performance over the following baselines:
\begin{itemize}
\item Random sampling- the robot selects a random action within its action space at each time step
\item Fixed sampling- When one sensor is involved, a lawnmower pattern is popular as it provides uniform coverage. When there are multiple sensors and a sensing budget involved, it is non trivial to design such paths. We here use a 5 stage policy which involves the robot using the remote sensor in the forward direction, 90 degrees to the left, and 90 degrees to the right, using the local sensor in the current cell and then moving one step forward. The stages are repeated until the robot's sensing budget is exhausted.
\item A greedy planner which selects the action with the highest immediate expected information gain to cost ratio. The behavior is similar to a frontier based strategy often used in exploration problems. The expected information gain is approximated by sampling observations from the belief space and simulating belief updates.
\end{itemize}

Large random environments were generated in which the location type and UV nodes were set to be a $40\times 40$ grid. Location type is the scientific latent variable of interest, which represents abstract geological features such as desert, riverbed, etc. The grid was further divided into 25 $8\times 8$ regions of homogeneous location types. The rock and feature space grids were of size $800 \times 800$. Each location grid cell therefore contains multiple rocks with associated features. The remote sensor can make observations in the feature space grid with a field of view of size 50 by 40 cells. All nodes were assigned ground truth labels by randomly sampling from the BN. An example environment is shown in Fig.~\ref{egmap}.  

The robot can occupy any one of the cells in the 40 by 40 grid and orient itself in 8 directions in $45\degree$ increments. In each decision step the robot can move one step forward in the direction it is facing or rotate on the spot with either -90$\degree$,-45$\degree$,45$\degree$ or 90$\degree$ increments. It also has to decide which of the two sensors to use. The size of the action space is therefore 10 actions. The $cost(a)$ function is defined as 1 unit for the remote sensor and 8 units for the local sensor. 

\begin{table}[!t]
\centering
\caption{Information gain with varying sensing budgets}
\label{table1}
\begin{tabular}{|c|c|c|c|}
\hline
                & \multicolumn{3}{c|}{\textbf{Sensing Budget}} \\ \hline
\textbf{Policy} & \textbf{50}   & \textbf{70}  & \textbf{100}  \\ \hline
Random          & $103.67(18.68)$ & $114.03(17.87)$ & $130.99(18.29)$               \\ \hline
Fixed           & $109.06(18.48)$ & $134.82(20.38)$ & $157.38(17.24)$              \\ \hline
Greedy          & $176.34(25.76)$ & $192.44(32.76)$ & $231.55(49.57)$               \\ \hline
MCTS-50         & $166.56(38.20)$ & $202.55(39.63)$ & $243.59(53.45)$               \\ \hline
MCTS-100        & $193.63(39.76)$ & $203.36(40.11)$ & $256.65(50.80)$               \\ \hline
\end{tabular}
\end{table}

\begin{table}[t]
\centering
\caption{Accuracy score with varying sensing budgets}
\label{table2}
\begin{tabular}{|c|c|c|c|}
\hline
                & \multicolumn{3}{c|}{\textbf{Sensing Budget}} \\ \hline
\textbf{Policy} & \textbf{50}   & \textbf{70}  & \textbf{100}  \\ \hline
Random          & $391.84(15.41)$ & $397.22(16.64)$ & $402.12(19.75)$              \\ \hline
Fixed           & $389.62(16.27)$ & $402.55(17.27)$ & $412.47(18.27)$             \\ \hline
Greedy          & $426.10(20.80)$ & $436.78(18.61)$ & $451.95(30.15)$             \\ \hline
MCTS-50         & $423.29(24.85)$ & $444.47(26.27)$ & $460.02(36.44)$           \\ \hline
MCTS-100        & $436.35(27.58)$ & $445.21(24.94)$ & $466.22(29.48)$             \\ \hline
\end{tabular}
\end{table}

We ran 50 trials for each policy with randomly generated environments and start locations. The policies were tested with sensing budgets of 50, 70 and 100 units. Two performance measures were used: the total information gained and an accuracy score. This is defined as the probability of the correct location class in the robot's belief. For example if a robot's belief about the class of $L$ in a particular cell is $[0.1,0.2,0.7]$ and the true class is the second one, the accuracy for the cell will be $0.2$. The accuracy score is the sum of the accuracy of all of the cells. It is an important metric because it captures situations in which the robot's belief converges to the wrong class. 

\begin{figure*}[!t]
\centering
\includegraphics[trim = {0 0.2 0 0.1cm},clip, width=\textwidth]{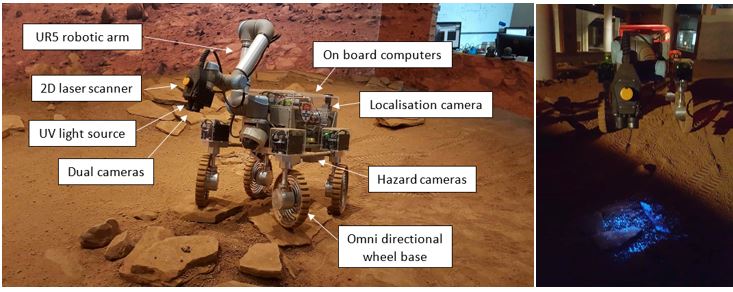}
\caption{Left: System diagram of Continuum. Right: Continuum's UV light source in action}
\label{continuum_diag}
\end{figure*} 

The average information gain and accuracy scores at the end of the mission are shown in Tables I and II with the standard deviation in brackets. Since our MCTS based planner is anytime, it was run with 50 and 100 iterations to test the effect of computation time on resulting performance. For all budget sizes, the adaptive algorithms (greedy and MCTS variants) significantly outperformed random and fixed sampling paths. For budget sizes 70 and 100 both of the MCTS variants yielded better performance than greedy in terms of both information gain and accuracy score. 

For a budget of 50 however, the greedy algorithm outperformed the MCTS-50 variant. We believe this is the case due to two reasons. Firstly, the simulation environment is open and unconstrained. With a small budget, the greedy strategy does not reach a point where the local information the robot can gain is exhausted. Secondly, in short planning horizons the next best action has a large effect on the final performance. Since the greedy algorithm allocated 20 samples for each action but MCTS-50 on average only uses 5 samples (the action space has a size of 10), the greedy approach has a better estimate of the information that can be gained in the next action. The fact that MCTS-100 significantly outperformed greedy supports this hypothesis.  

In terms of computation time, each iteration of the MCTS took between 0.2 to 0.5 seconds on an average desktop computer. The implementation was however in MATLAB and can be significantly sped up through more efficient memory management and handling of data structures. Parts of the algorithm can be parallelized so utilizing multi-threading is also a possibility.

\section{Planetary Rover Experiments}

In this section we demonstrate the practicality of our approach with a rover mission on an analog Martian terrain based in the Museum of Applied Arts and Sciences (MAAS) in Sydney. This section summarizes the platform capabilities, testing environment, our computer vision technique and concludes with some trial experiments.

\subsection{Platform Details}
Our rover Continuum is pictured in Fig. \ref{continuum_diag}. It is equipped with an omni-directional drive which gives it relatively unconstrained motion capabilities. The spiral shape of the rims act as shock absorbers while the double-bogie chassis allow the rover to climb over steep rocks and minimize the changes in orientation. Continuum has a 6 degree of freedom robotic arm with cameras, an ultraviolet light source and a 2D laser scanner mounted on the end effector. There are also several hazard cameras around the body to check for collisions. In this experiment we use one of the arm cameras and the UV light source as our two sensors. The light source illuminates the UV reflective powder we discuss in the next section and simulates what a spectrometer might do in a real mission. The arm camera was pointed towards the ground to constrain the information that can be gathered in a single sense.

\begin{figure*}[t]
\centering
\includegraphics[width=\textwidth]{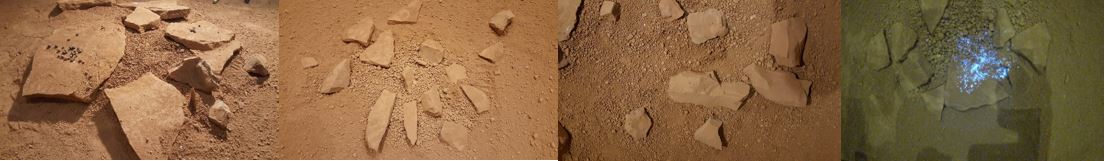}
\caption{From left to right: The three classes of location type and a typical image when the local sensor is activated.}
\label{area_example}
\end{figure*}

\begin{figure*}[t]
\centering
\includegraphics[width=\textwidth]{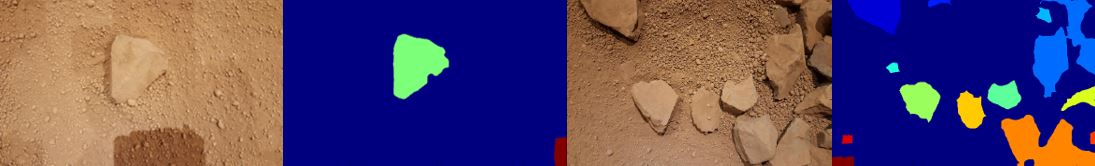}
\caption{Our rock segmentation technique in action. It can be seen that there are still some false positives in areas with shadows.}
\label{cv_example}
\end{figure*}

\subsection{Environment Setup}
Our testing environment, the MAAS Mars Lab is a $20\times 7$m space which is designed to be a scientifically accurate representation of Martian terrain. The lab was divided into three different types of location shown in Fig. \ref{area_example}. Each location type had slightly different distributions of types of rocks and the features they exhibit. UV reflective powder was added in varying quantities to each category. There was however enough ambiguity between categories to encourage the robot to use a combination of both sensors to gather information. The rock grid was set to be a resolution of 2cm per cell. Rocks are different sizes so they usually span across many cells. To account for this we assume they are located in the cell nearest to their centroid. The conditional probability parameters of the BN were determined from intuition and therefore not 100\% accurate. There were also rocks in the environment which were not explicitly modeled in the BN, which is a realistic source of noise not present in the simulations. 

\subsection{Computer vision}
In a realistic unstructured environment the feature extraction process is more complex and requires first segmenting the rocks from the image.  It can be seen in Figures 6, 7 and 8 that rocks look very similar to ground in terms of colour. There are also lighting variations and shadows which complicate the image processing step. There are several methods proposed in literature which achieved good results. Edge-based techniques such as \cite{thompson2007performance} ran a Canny edge detector followed by a complex process of pruning and joining edges likely to belong to a rock. Texture based techniques such as \cite{song2008automated} utilized multi-resolution histograms to achieve coarse segmentation followed by an active contour technique to get good edge detection performance. Another interesting and effective approach was used by \cite{dunlop2007multi} which calculated superpixels at different scales followed by adding, subtracting, splitting and merging superpixels to satisfy criterion learned from a Support Vector Machine. However all of these approaches were designed for Martian imagery which did not have the same characteristics as our environment and were not available open source. Furthermore computation time was not considered in these studies so the algorithms often took several minutes to yield a result. 

We approach this problem by first over-segmenting the image into superpixels using the SLIC algorithm \cite{achanta2012slic} which groups similarly colored pixels together while preserving the strong edges. This is followed by adaptive normalization to reduce lighting variations and shadows. Histograms of intensity, the number of edges, LAB color and intensity variance were calculated for each superpixel and compared to a training image of the ground with no rocks. Applying appropriate thresholds allows us to classify most of the superpixels as rock, ground or shadow. For the more uncertain superpixels, the amount of texture correlation with their local neighborhoods was measured followed by a voting process. This two stage process yields the final image shown in Fig. \ref{cv_example}. Segmentation is sometimes noisy like most robotic applications especially in the presence of shadows but the probabilistic nature of Bayesian networks helps minimize the resulting effects on decision making. For features we use circularity, size and color as they are simple to calculate and geologically meaningful. The UV measurement was obtained by calculating the blue to red ratio of the RGB channels. The features and UV measurements were both discretized into three categories.

\subsection{Localization and control}
PID controllers were used in conjunction with a localization system detailed in previous work \cite{potiris2014terrain} to control the omni-directional drive such that the required position and orientation is achieved within a small error margin. Localization was fused with the computer vision to register observations on a map which allowed the belief space to be updated. The action space was once again discretized into ten actions where the robot could select one of two sensors and decide whether to move forward one step, move diagonally at -45 and 45 degrees or rotate by -90 or 90 degrees. The robot also checked if actions will lead to collisions or cause the robot to drive over valuable rocks through an occupancy map provided to the robot prior to the mission. 


\subsection{Results} 
We compared our non-myopic planner against a random action policy with random start locations and orientations in the yard. Ten trials were run for each policy. A sensing budget of 30 units was used with a cost function of 1 and 5 units for the remote and local sensor respectively. We also attempted to implement a greedy strategy but found early in the trials that the robot often got stuck in local minima and wasn't able to give useful results. This is because, unlike the simulations, there were many non traversable rocks present which often created concave areas in the occupancy map. A random policy was able to better recover from such situations, and hence was a better benchmark to compare our algorithm against. The information gain and accuracy scores along with standard deviations are shown in Table III. 

\begin{table}[!t]
\centering
\caption{Performance comparison of MCTS planner with random for real robot experiments}
\label{my-label}
\begin{tabular}{|c|c|c|ll}
\hline
\textbf{Policy} & \textbf{Information Gain} & \textbf{Accuracy Score}   \\ \hline
Random    & 52.23 (11.76)  & 161.89 (8.48)   \\ \hline
MCTS-50   & 59.17 (18.63)  & 170.04 (10.66)   \\ \hline
\end{tabular}
\end{table}

If the robot has a uniform distribution over the belief of $L$ across all the cells, the accuracy score is $139.33$. The MCTS algorithms therefore gives almost a $25\%$ increase in accuracy score over random policies and $13\%$ increase in terms of information gain. It is important to note the testing environment was relatively small. Longer horizon plans are likely to generate even more performance benefits. 

\section{Conclusions and Future Work}
The results presented in this paper show that our approach has the potential to extend the autonomy of space rovers, and information gathering robots in general. A novel method for encoding scientific knowledge in a BN was proposed, along with a MCTS planner to reason about the network and create informative action policies. This enables robots to plan and deploy sensors to directly study scientifically interesting latent variables in a closed loop fashion. The reduced reliance on communication with scientists for navigation should lead to increased science returns in future missions. Our approach was tested extensively in simulation as well as in an analog Mars environment and showed significant performance improvements over simpler policies. 

In future work we would like to evaluate our approach in different use cases such as agriculture and remote sensing. Richer knowledge representation frameworks such as statistical relational models could be explored, while the performance of the MCTS can be further improved through more informed rollout policies and better reward function approximations. Another interesting line of work is to adapt the structure and conditional probability parameters of the BN online to better fit and predict observations.  

\section*{ACKNOWLEDGMENT}
We would like to thank ACFR, MAAS and the Mars Lab project for supporting this work. Thanks also goes to Graeme Best, Oliver Cliff, Asher Bender and Steven Potiris for their valuable feedback. 


\bibliographystyle{ieeetr}
\bibliography{biblio_IROS}

\end{document}